# Line of Sight Curvature for Missile Guidance using Reinforcement Meta-Learning

Brian Gaudet*
*University of Arizona, 1127 E. Roger Way, Tucson Arizona, 85721*

Roberto Furfaro†
*University of Arizona, 1127 E. Roger Way, Tucson Arizona, 85721*

**We use reinforcement meta learning to optimize a line of sight curvature policy that increases the effectiveness of a guidance system against maneuvering targets. The policy is implemented as a recurrent neural network that maps navigation system outputs to a Euler 321 attitude representation. The attitude representation is then used to construct a direction cosine matrix that biases the observed line of sight vector. The line of sight rotation rate derived from the biased line of sight is then mapped to a commanded acceleration by the guidance system. By varying the bias as a function of navigation system outputs, the policy enhances accuracy against highly maneuvering targets. Importantly, our method does not require an estimate of target acceleration. In our experiments, we demonstrate that when our method is combined with proportional navigation, the system significantly outperforms augmented proportional navigation with perfect knowledge of target acceleration, achieving improved accuracy with less control effort against a wide range of target maneuvers.**

## I. Introduction

Proportional navigation (PN) can be shown to be equivalent to a guidance law that minimizes the zero effort miss (ZEM) [1], with the ZEM being the predicted miss distance in the case that neither the missile nor the target accelerates for the remainder of the engagement. Consequently, target maneuvers can increase miss distance against a missile employing PN guidance. However, if the target acceleration can be estimated, the PN guidance law can be augmented using an additional term that is a function of estimated target acceleration, allowing improved performance against maneuvering targets. This augmented proportional navigation (APN) guidance law can be shown to be optimal for targets employing a step acceleration maneuver [2] and can increase accuracy for other target maneuvers as well. Other formulations of PN give improved performance against weaving targets, provided the missile navigation system can estimate the weave frequency. However, accurate estimation of target acceleration is difficult for arbitrary target maneuvers, and there is the possibility that the estimate can diverge, resulting in large miss distances [3]. For this reason, there has been considerable research effort devoted to guidance laws that perform well against maneuvering targets, but do not require estimates of target acceleration.

In [4] the authors augment true proportional navigation [5] with an additional term that is the product of a time-varying constant and the estimated target acceleration, with the additional term derived through capturability analysis. However, the experiments only tested the guidance law with a relatively large ratio of missile to target acceleration capability ( 50:1 in Figure 12). In another approach [6] a guidance law that does not require an estimate of target acceleration is developed, and the authors show that the guidance law requires less acceleration than PN for a single target maneuver, but do not demonstrate effectiveness over a range of target maneuvers. Switched bias proportional navigation (SBPN), described in [3], uses sliding mode theory to derive a bias term that is added to the commanded acceleration from PN, with the bias term only depending on the sign of the line of sight rotation rate. SBPN results in an acceleration profile close to that of APN, without requiring an estimate of target acceleration. Other proposed guidance laws use geometric guidance [7, 8], parameterizing the missile trajectory in the Fresnet-Serret reference frame. Importantly, [7] assumes knowledge of the target acceleration vector, and [8] only considers low speed targets.

---

*Research Engineer, Department of Systems and Industrial Engineering, E-mail: briangaudet@arizona.edu
†Professor, Department of Systems and Industrial Engineering, Department of Aerospace and Mechanical Engineering. E-mail: robertof@arizona.edu

In contrast, we propose a new approach to improving the performance of any guidance law against maneuvering targets. Specifically, we optimize a line of sight curvature policy $\pi$ that maps navigation system outputs to a Euler 321 attitude parameterization $\theta_{\text{LOSC}}$. $\theta_{\text{LOSC}}$ is then mapped to a direction cosine matrix, which rotates the observed line of sight (LOS) unit vector pointing from the missile to the target. By varying $\theta_{\text{LOSC}}$ as a function of navigation system outputs, $\pi$ can arbitrarily curve the LOS during an engagement. We parameterize $\pi$ as a deep neural network with recurrent layer optimized using reinforcement meta-learning (meta-RL) [9].

RL has been demonstrated to be effective in optimizing integrated and adaptive G&C systems that generate direct closed-loop mapping from navigation system outputs to actuation commands. Applications where RL has been effectively applied to GN&C include asteroid close proximity operations [10, 11], planetary landing [12–14], exoatmospheric intercept [15], endoatmospheric intercept [16], and hypersonic vehicle guidance [17, 18]. In the RL framework, an agent instantiating a policy learns how to complete a task through episodic simulated experience with an environment. The policy is implemented as a deep neural network that maps observations to actions $\mathbf{u} = \pi_\theta(\mathbf{o})$, and in our work is optimized using a customized version of proximal policy optimization (PPO)[19]. In this work we include a recurrent layer in the policy and value function networks. This allows making the LOS curvature a function of the history of observations $\mathbf{o}$.

We optimize the LOS curvature policy in an environment where the curved LOS is passed to a proportional navigation guidance system, creating the PN-LOSC guidance law. We test PN-LOSC over a wide range of target maneuvers, a large altitude range, and using a simple aerodynamic drag model for the missile and target. Missile and target maximum acceleration are a function of dynamic pressure, which would be the case for vehicles maneuvering using control surface deflections. In our experiments we demonstrate that PN-LOSC achieves improved accuracy with less control as compared to both PN and APN against maneuvering targets with an acceleration capability approaching that of the missile. Importantly, although we demonstrate the LOS curvature approach using PN guidance, a LOS curvature policy can be optimized in conjunction with any guidance law that takes inputs derived from the LOS unit vector pointing from the missile to the target.

The paper is organized as follows. In Section II we present the engagement scenarios, LOS curvature model, guidance system model, and the equations of motion. Next in Section III.A we give a brief summary of the RL framework. This is followed by Section III.B, where we formulate the RL optimization problem for the air-to-air homing phase scenario described in Section II. In Section IV we optimize and test the PN-LOSC policy, and compare the results to PN and APN benchmarks.

## II. Problem Formulation

### A. Engagement Geometry and Initial Conditions

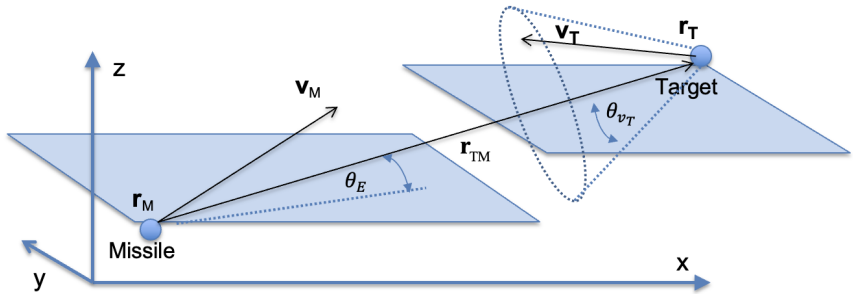

**Fig. 1  Engagement**

We consider a skewed head on engagement scenario. Due to the high target acceleration capability, which exceeds the level that a human pilot could withstand, a realistic scenario would be a fighter launching a missile to intercept a supersonic cruise missile that can actively maneuver to avoid threats. Referring to Fig. 1[*], the missile position vector, missile velocity vector, target position vector, and target velocity vector are shown as $\mathbf{r}_M$, $\mathbf{v}_M$, $\mathbf{r}_T$, $\mathbf{v}_T$. We can also define the relative position and velocity vectors $\mathbf{r}_{TM} = \mathbf{r}_T - \mathbf{r}_M$ and $\mathbf{v}_{TM} = \mathbf{v}_T - \mathbf{v}_M$. The elevation angle $\theta_E$ is the angle between $\mathbf{r}_{TM}$ and its projection onto the x-y plane. We randomly generate the target's initial velocity vector such that $\mathbf{v}_T$ lies within a cone with axis $\mathbf{r}_{TM}$ and half apex angle $\theta_{\mathbf{v}_T}$. A collision triangle is then defined in a plane that is not in general aligned with the coordinate frame shown in Fig. 1, and is illustrated in Fig. 2. Here we define the required lead angle $L$ for the missile's velocity vector $\mathbf{v}_M$ as the angle that will put the missile on a collision triangle with the target in terms of the target velocity $\mathbf{v}_T$, line-of-sight angle $\gamma$, and the magnitude of the missile velocity as shown in Eqs. (1a) through (1c).

---
[*]In this figure, the illustrated vectors are not in general within the x-z plane

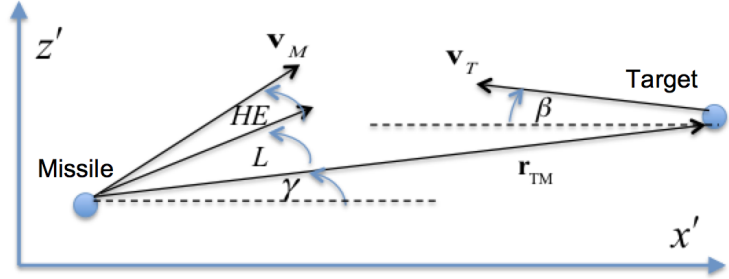

**Fig. 2  Planar Heading Error**

$$L = \arcsin\left(\frac{\|\mathbf{v}_T\| \sin(\beta + \gamma)}{\|\mathbf{v}_M\|}\right) \tag{1a}$$

$$v_{m_y} = \|\mathbf{v}_M\| \cos(L + \gamma) \tag{1b}$$

$$v_{m_z} = \|\mathbf{v}_M\| \sin(L + \gamma) \tag{1c}$$

This formulation is easily extended to a three dimensional engagement using the following approach:
1) define a plane normal as $\hat{\mathbf{v}}_t \times \hat{\boldsymbol{\lambda}}$
2) rotate $\mathbf{v}_T$ and $\hat{\boldsymbol{\lambda}}$ onto the plane
3) calculate the required planar missile velocity (Eq. (1a))
4) rotate this velocity back into the original reference frame

Thus in $\mathbb{R}^3$ we define a heading error (HE) as the angle between the missile's initial velocity vector and the velocity vector associated with the lead angle required to put the missile on a collision heading with the target. Note that due to the missile aerodynamic forces and target acceleration, this is far from a perfect collision triangle, and the true heading error is generally greater than HE.

The simulator randomly chooses between a target bang-bang, weave, and jinking maneuver with equal probability, with the acceleration applied orthogonal to the target's velocity vector. The maneuvers have varying acceleration levels and random start time, duration, and switching time. At the start of each episode, with probability 0.5 the maneuvers in that episode use the target's maximum acceleration capability, and with probability 0.5 the acceleration is sampled uniformly between 0 and the maximum. We assume the target uses aerodynamic control surfaces (no thrust vector control). Consequently, the maximum target acceleration is reduced taking into account dynamic pressure. Specifically, we assume that the target can achieve the acceleration shown in Table 1 only at $q_o^{\text{MAX}}$, the dynamic pressure corresponding to its maximum speed at sea level, but we reduce this maximum acceleration by the ratio of $\dfrac{q_o^{\text{MAX}}}{q_o}$, where $q_o = \dfrac{1}{2}\rho\|\mathbf{v}_T\|^2$. Sample target maneuvers are shown Fig. 3, note that in some cases the maneuver period is considerably shorter or longer, with the longest periods being twice the time of flight.

We can now list the range of engagement scenario parameters in Table 1. During optimization and testing, these parameters are drawn uniformly between their minimum and maximum values, except as noted. The generation of heading error is handled as follows. We first calculate the optimal missile velocity vector that puts the missile on a collision triangle with the target as described previously. We then uniformly select a heading error $HE$ between the bounds given in Table 1, and randomly perturb the direction of the missile's velocity vector direction such that $\arccos(\mathbf{v}_M \cdot \mathbf{v}_{M_p}) < HE$, where $\mathbf{v}_{M_p}$ is the perturbed missile velocity vector.

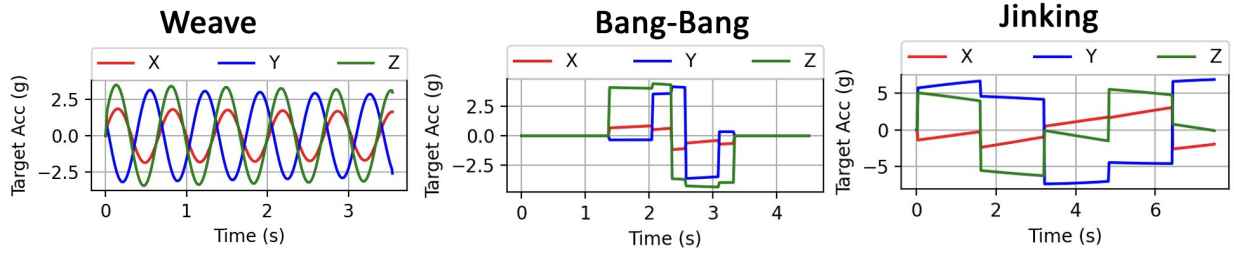

Fig. 3    Sample Target Maneuvers

Table 1    Simulator Initial Conditions for Optimization

| Parameters Drawn Uniformly | Min | Max |
| --- | --- | --- |
| Range $\|\mathbf{r}_{TM}\|$ (m) | 5000 | 10000 |
| Elevation Angle $\theta_E$ (degrees) | -30 | 30 |
| Missile Velocity Magnitude $\|\mathbf{v}_M\|$ (m/s) | 800 | 1000 |
| Missile Maximum Acceleration @ $q_o^{MAX}$ (m/s$^2$) | 40 | 40×9.81 |
| Target Velocity Magnitude $\|\mathbf{v}_T\|$ (m/s) | 400 | 600 |
| Target Velocity Cone Half Apex Angle $\theta_{\mathbf{v}_T}$ (degrees) | 30 | 30 |
| Heading Error (degrees) | 0 | 5 |
| Target Maximum Acceleration @ $q_o^{MAX}$ (m/s$^2$) | 0 | 30×9.81 |
| Target Bang-Bang duration (s) | 1 | 8 |
| Target Bang-Bang initiation time (s) | 0 | 6 |
| Target Weave Period (s) | 1 | 8 |

### B. Radome and Seeker Model

We use a similar radome model to that used in [16], but simplified for the 3-DOF simulator used in this work. Here the look angle is defined as the angle between the ground truth inertial frame line of sight (LOS) vector $\lambda = \frac{\mathbf{r}_{TM}}{\|\mathbf{r}_{TM}\|}$ and the missile's velocity unit vector $\hat{\mathbf{v}}_M = \frac{\mathbf{v}_M}{\|\mathbf{v}_M\|}$, with the look angle calculated as $\theta_L = \arccos(\lambda \cdot \hat{\mathbf{v}_{TM}})$. This definition of look angle is different than in a 6-DOF environment, where the look angle is the angle between the missile centerline axis and $\lambda$. However, the dynamic look angle in our 3-DOF implementation coupled with our look angle dependent refraction model does create a parasitic attitude loop [20], reducing guidance system performance.

The radome refraction angle $\theta_R$ is defined as the angle between the ground truth LOS $\lambda$ and the apparent LOS $\tilde{\lambda}$, i.e., $\theta_R = \arccos(\lambda \cdot \tilde{\lambda})$. We assume a symmetrical radome, where the radome refraction angle $\theta_R$ is a function of look angle $\theta_L$. The model first calculates the azimuthal ($\theta_u$) and elevation ($\theta_v$) refraction errors as shown in Eqs. (3a) through (3b), where $A_u$, $A_v$, $k_u$, and $k_v$ are sampled uniformly within the bounds given in Table 2 at the start of each simulation episode. We then create the refracted body frame LOS unit vector as shown in Eq. (2a), where $\mathbf{C}(\mathbf{q}_R)$ is the DCM corresponding to the 321 Euler rotation $\mathbf{q}_R = [\theta_u, \theta_v, 0]$. $\mathbf{q}_N = \mathcal{N}(\mu, \sigma_{LOS}, 3)$ is a stochastic Euler 321 rotation (to model Gaussian white noise on the LOS measurement), and $\mathcal{N}(\mu, \sigma, n)$ denotes an $n$ dimensional normally distributed random variable with mean $\mu$ and standard deviation $\sigma$. We use $\sigma_{LOS} = 1$ mrad. "LAG" indicates a first order low pass noise filter with time constant 0.02s. We assume a strap down seeker implementation, and therefore do not model seeker gimbal lag.

$$\tilde{\lambda} = \text{LAG}\bigg(\mathbf{C}(\mathbf{q}_N)\big(\mathbf{C}(\mathbf{q}_R)\lambda\big)\bigg) \tag{2a}$$

A plot of refraction angle $\theta_R$ as a function of the look angle $\theta_L$ is shown in Fig. 4 for the case of $A_u = A_v = 10$ mrad and various values of $k$. Note that the radome slope $\frac{\partial \theta_R}{\partial \theta_L}$ is given by the slope of the curves in the figure.

The simplified navigation system outputs the relative range $r = \|\mathbf{r}_{TM}\|$, closing velocity $v_c = -\tilde{\boldsymbol{\lambda}} \cdot \mathbf{v}_{TM}$, the ground truth relative velocity vector $\mathbf{v}_{TM}$, and the LOS rotation rate $\tilde{\boldsymbol{\Omega}} = \dfrac{\tilde{\mathbf{r}}_{TM} \times \mathbf{v}_{TM}}{\tilde{\mathbf{r}}_{TM} \cdot \tilde{\mathbf{r}}_{TM}}$, where $\tilde{\mathbf{r}}_{TM} = \tilde{\boldsymbol{\lambda}} r$

$$\theta_u = A_u \left( 0.75 \frac{\theta_L}{\pi/2} + 0.25 \cos\left( \frac{2\pi}{k_u} \theta_L \right) \right) \tag{3a}$$

$$\theta_v = A_v \left( 0.75 \frac{\theta_L}{\pi/2} + 0.25 \cos\left( \frac{2\pi}{k_v} \theta_L \right) \right) \tag{3b}$$

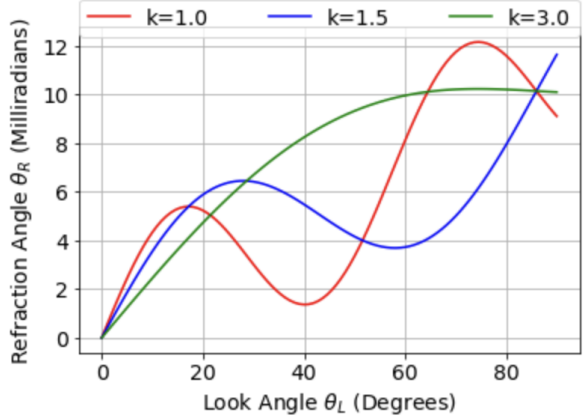

**Fig. 4  Radome Refraction Angle as Function of Look Angle**

**Table 2  Radome Model Parameter Bounds**

| Variable | Lower limit | Upper Limit |
|---|---|---|
| $A_u$ | -1e-2 | 1e-2 |
| $A_v$ | -1e-2 | 1e-2 |
| $k_u$ | 1.00 | 3.00 |
| $k_v$ | 1.00 | 3.00 |

## C. LOS Curvature

The LOS curvature policy $\pi : \mathbf{o} \mapsto \mathbf{u}$ maps observations $\mathbf{o}$ to actions $\mathbf{u}$, with the observations given in Eq. (4).

$$\mathbf{o} = \begin{bmatrix} \tilde{\boldsymbol{\lambda}} & \tilde{\boldsymbol{\Omega}} & v_c & r \end{bmatrix} \tag{4}$$

The action $\mathbf{u}$ is then scaled and interpreted as a Euler 321 attitude, as shown in Eq. (5), where we use $k = 2°$.

$$\boldsymbol{\theta}_{\text{LOSC}} \in SO(3) = k\mathbf{u} \tag{5}$$

$\boldsymbol{\theta}_{\text{LOSC}}$ is then used to construct the direction cosine matrix (DCM) $\mathbf{C}(\boldsymbol{\theta}_{\text{LOSC}})$, and compute the shaped LOS direction vector as $\boldsymbol{\lambda}_{\text{LOSC}} = \mathbf{C}(\boldsymbol{\theta}_{\text{LOSC}})\tilde{\boldsymbol{\lambda}}$. Thus, by varying $\boldsymbol{\theta}_{\text{LOSC}}$ during the engagement, the LOS curvature policy can arbitrarily curve $\boldsymbol{\lambda}_{\text{LOSC}}$. The signal $\boldsymbol{\lambda}_{\text{LOSC}}$ is then used to construct a LOS rotation vector $\boldsymbol{\Omega}_{\text{LOSC}} = \dfrac{\mathbf{r}_{TM_{\text{LOSC}}} \times \mathbf{v}_{TM}}{\mathbf{r}_{TM_{\text{LOSC}}} \cdot \mathbf{r}_{TM_{\text{LOSC}}}}$, where $\mathbf{r}_{TM_{\text{LOSC}}} = \boldsymbol{\lambda}_{\text{LOSC}} r$. Both $\boldsymbol{\Omega}_{\text{LOSC}}$ and $\boldsymbol{\lambda}_{\text{LOSC}}$ are used by the guidance law described in Section II.D.

$\pi$ is implemented as a deep recurrent neural network and optimized using reinforcement meta-learning, as described in Sections III.A and III.B. Both $\pi$ and the value function used to optimize $\pi$ contain a recurrent network layer, allowing actions to be generated using the history of observations. In theory, this allows the $\pi$ to infer properties of the target maneuvers.

## D. Guidance Law

The guidance law implements the true proportional navigation (TPN) guidance law [21], which is shown in Eqs. (6a), with the commanded acceleration adjusted so that it is perpendicular to $\mathbf{v}_M$ in Eqs. (6b) and (6c), where $\hat{\mathbf{r}}_{TM} = \frac{\mathbf{r}_{TM}}{\|\mathbf{r}_{TM}\|}$, and $\mathbf{\Omega}$ is the LOS rotation rate [21].

$$\mathbf{a} = -N v_c (\lambda_{\text{LOSC}} \times \mathbf{\Omega}_{\text{LOSC}}) \tag{6a}$$

$$\mathbf{a}_{\text{PAR}} = \mathbf{a} \cdot \mathbf{v}_{TM} \frac{\mathbf{v}_{TM}}{\|\mathbf{v}_{TM}\|} \tag{6b}$$

$$\mathbf{a}_{\text{COM}} = \mathbf{a} - \mathbf{a}_{\text{PAR}} \tag{6c}$$

The APN benchmark uses the TPN guidance law described in Eqs. (6a) and (6b), except that the APN benchmark replaces Eq. (6a) with Eq. (7a).

$$\mathbf{a} = -N v_c (\lambda_{\text{LOSC}} \times \mathbf{\Omega}_{\text{LOSC}}) + N \frac{\mathbf{a}_T}{2} \tag{7a}$$

In Section IV the benchmark PN and APN guidance laws use $k = 0$ in Eq. (5), i.e., there is no LOS curvature. With $k = 2°$, we obtain the PN-LOSC guidance law. A system diagram of PN-LOSC guidance is illustrated in Fig. 5. The simulation environment used to optimize and test the PN, APN, and PN-LOSC guidance laws instantiates this system. We use $N = 3$ for both PN and APN guidance, noting that this is optimal for APN.

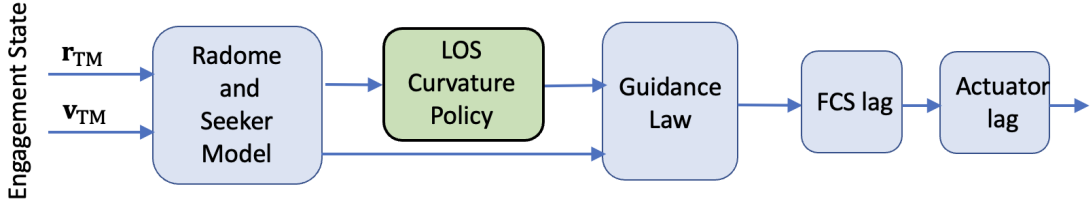

**Fig. 5  System Diagram**

Eqs. 8a and 8b model the lag from the flight control system (FCS) and actuator. Here we define $a_{M_{\text{REF}}}$ as the magnitude of the maximum achievable lateral acceleration at sea level and 1000 m/s. We set $a_{M_{\text{REF}}} = 74g$, which corresponds to the steady state acceleration at those conditions with a 20 degree horizontal and vertical fin deflection in our 6-DOF simulator [16], and $g = 9.81$ m/s$^2$. $a_M$ is then clipped based off of dynamic pressure, where $\rho(h)$ is the atmospheric density at missile altitude $h_T$ and $V = \|\mathbf{v}_M\|$, and again clipped to remain below the load constraint of $a_{M_{\text{MAX}}} = 40g$. In Eq. (8a), "F-LAG" denotes a first order low pass filter with time constant 0.08s that we use to represent the flight control system time constant, and "A-LAG" denotes a first order low pass filter with time constant of 0.02s that is used to model actuator lag.

$$a_M = \text{F-LAG}\left(\text{clip}\left(\text{clip}\left(\|\mathbf{a}_{\text{COM}}\|, 0, \frac{\rho(h_M) V_M^2}{\rho(0) 1000^2} a_{M_{\text{REF}}}\right), 0, a_{M_{\text{MAX}}}\right)\right) \tag{8a}$$

$$\mathbf{a}_M = \text{A-LAG}\left(a_M \frac{\mathbf{a}_{M_{\text{COM}}}}{\|\mathbf{a}_{M_{\text{COM}}}\|}\right) \tag{8b}$$

## E. Equations of Motion

The missile inertial frame position $\mathbf{r}_M$ and velocity unadjusted for drag $\tilde{\mathbf{v}}_M$ are then updated by integrating Eqs. (9a) and (9b). The missile speed $V_M = \|\mathbf{v}_M\|$ is adjusted for drag by integrating Eq. (9c), where we use $k_M = 1/4$, $m_M = 450$ kg, cd0$_M = 0.35$. $q_M = \frac{h \mathbf{v}_M^2}{2}$ is the missile dynamic pressure.

$$\dot{\mathbf{r}}_M = \mathbf{v}_M \tag{9a}$$
$$\dot{\tilde{\mathbf{v}}}_M = \mathbf{a}_M \tag{9b}$$
$$\dot{V}_M = -\frac{q_M \text{cd0}_M}{m_M} - k_M \|[\mathbf{a}_M]\| \tag{9c}$$

The missile velocity is then computed as shown in Eq. (10)

$$\mathbf{v}_M = V_M \frac{\tilde{\mathbf{v}}_M}{\|\tilde{\mathbf{v}}_M\|} \tag{10}$$

The target is modeled as shown in Eqs. (11a) through (11b), where $\mathbf{a}_{T_{\text{COM}}}$ is the commanded target acceleration assuming the maneuvers described in Section II.A. The target maximum speed is 600m/s, which appears in the denominator of Eq. (11a).

$$a_T = \|\mathbf{a}_{T_{\text{COM}}}\| \frac{\rho(h_T) V_T^2}{\rho(0) 600^2} \tag{11a}$$
$$\mathbf{a}_T = a_T \frac{\mathbf{a}_{T_{\text{COM}}}}{\|\mathbf{a}_{T_{\text{COM}}}\|} \tag{11b}$$

The target inertial frame position $\mathbf{r}_T$ and velocity unadjusted for drag $\tilde{\mathbf{v}}_T$ are then updated by integrating Eqs. (12a) and (12b). The target speed $V_T = \|\mathbf{v}_T\|$ is adjusted for drag by integrating Eq. (12c), where we use ranges of $k_T$ from 1/8 to 1/3 and ranges of $\text{cd0}_T$ from 0.125 to 0.4 for the experiments that randomize target drag; these parameters are both set to zero for the experiments where we do not model target drag. We use $m_T = 450$ kg.

$$\dot{\mathbf{r}}_T = \mathbf{v}_T \tag{12a}$$
$$\dot{\tilde{\mathbf{v}}}_T = \mathbf{a}_T \tag{12b}$$
$$\dot{V}_T = -\frac{\rho(h_T) V_T^2 \text{cd0}_T}{2 m_T} - k_T \|[\mathbf{a}_T]\| \tag{12c}$$

The target velocity is then computed as shown in Eq. (13)

$$\mathbf{v}_T = V_T \frac{\tilde{\mathbf{v}}_T}{\|\tilde{\mathbf{v}}_T\|} \tag{13}$$

The equations of motion are updated using fourth order Runge-Kutta integration. For ranges greater than 80 m, a timestep of 20 ms is used, and for the final 80 m of homing, a timestep of 0.2 ms is used in order to more accurately (within 0.4m) measure miss distance; this technique is borrowed from [22]. The policy and guidance system inputs are updated every 20 ms.

## III. Methods

### A. Reinforcement Learning Framework

In the reinforcement learning framework, an agent learns through episodic interaction with an environment how to successfully complete a task using a policy that maps observations **o** to actions **u**. The environment initializes an episode by randomly generating a ground truth state, mapping this state **x** to an observation, and passing the observation to the agent. The agent uses this observation to generate an action that is sent to the environment; the environment then uses the action and the current ground truth state to generate the next state and a scalar reward signal $r(\mathbf{x}, \mathbf{u})$. The reward and the observation corresponding to the next state are then passed to the agent. The process repeats until the environment terminates the episode, with the termination signaled to the agent via a done signal. Trajectories collected over a set of episodes (referred to as rollouts) are collected during interaction between the agent and environment, and used to update the policy and value functions. The interface between agent and environment is depicted in Fig. 6.

Meta-RL differs from generic reinforcement learning in that the agent learns over an ensemble of environments. These environments vary the engagement scenarios, dynamics, aerodynamic coefficients, radome parameters, and other factors. Optimization within the meta-RL framework results in an agent that can quickly adapt to novel environments, often with just a few steps of interaction with the environment. Similar to [23], we implement meta-RL by including a recurrent layer in the policy and value function. For a given trajectory over observations and actions, the recurrent layer will evolve differently for different target maneuvers, allowing the policy to infer the nature of the maneuver. By optimizing over an ensemble of target behavior models, the agent learns to adapt, using the recurrent layer's hidden state to infer the current target behavior model.

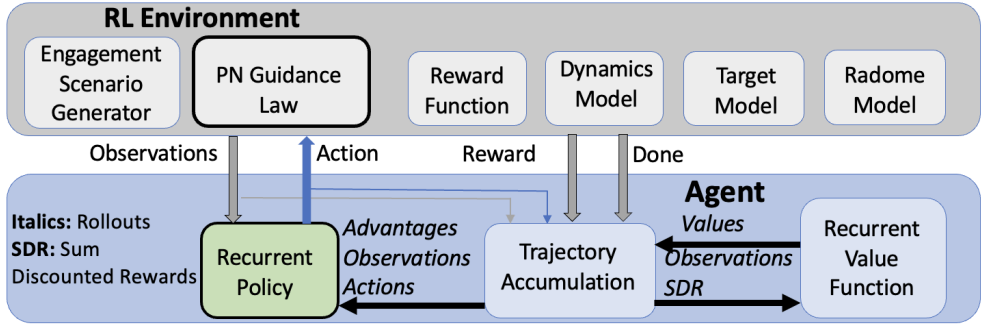

**Fig. 6  Environment-Agent Interface**

In this work, we implement RL using proximal policy optimization (PPO) [19] with both the policy and value function implementing recurrent layers in their networks. The PPO algorithm has demonstrated state-of-the-art performance for many reinforcement learning benchmark problems. PPO approximates the Trust Region Policy Optimization method [24] by accounting for the policy adjustment constraint with a clipped objective function. The objective function used with PPO can be expressed in terms of the probability ratio $p_k(\boldsymbol{\theta})$ given by,

$$p_k(\boldsymbol{\theta}) = \frac{\pi_{\boldsymbol{\theta}}(\mathbf{u}_k|\mathbf{o}_k)}{\pi_{\boldsymbol{\theta}_{\text{old}}}(\mathbf{u}_k|\mathbf{o}_k)} \tag{14}$$

The PPO objective function is shown in Equations (15a) through (15c). The general idea is to create two surrogate objectives, the first being the probability ratio $p_k(\boldsymbol{\theta})$ multiplied by the advantages $A_{\mathbf{w}}^{\pi}(\mathbf{o}_k, \mathbf{u}_k)$ (see Eq. (16)), and the second a clipped (using clipping parameter $\epsilon$) version of $p_k(\boldsymbol{\theta})$ multiplied by $A_{\mathbf{w}}^{\pi}(\mathbf{o}_k, \mathbf{u}_k)$. The objective to be maximized $J(\boldsymbol{\theta})$ is then the expectation under the trajectories induced by the policy of the lesser of these two surrogate objectives.

$$\text{obj1} = p_k(\boldsymbol{\theta}) A_{\mathbf{w}}^{\pi}(\mathbf{o}_k, \mathbf{u}_k) \tag{15a}$$

$$\text{obj2} = \text{clip}(p_k(\boldsymbol{\theta}) A_{\mathbf{w}}^{\pi}(\mathbf{o}_k, \mathbf{u}_k), 1 - \epsilon, 1 + \epsilon) \tag{15b}$$

$$J(\boldsymbol{\theta}) = \mathbb{E}_{p(\tau)}[\min(\text{obj1}, \text{obj2})] \tag{15c}$$

This clipped objective function has been shown to maintain a bounded Kullback-Leibler (KL) divergence [25] with respect to the policy distributions between updates, which aids convergence by ensuring that the policy does not change drastically between updates. Our implementation of PPO uses an approximation to the advantage function that is the difference between the empirical return and a state value function baseline, as shown in Equation 16, where $\gamma$ is a discount rate and $r$ the reward function.

$$A_{\mathbf{w}}^{\pi}(\mathbf{x}_k, \mathbf{u}_k) = \left[\sum_{\ell=k}^{T} \gamma^{\ell-k} r(\mathbf{x}_\ell, \mathbf{u}_\ell)\right] - V_{\mathbf{w}}^{\pi}(\mathbf{x}_k) \tag{16}$$

Here the value function $V_{\mathbf{w}}^{\pi}$ is learned using the cost function given by

$$L(\mathbf{w}) = \frac{1}{2M} \sum_{i=1}^{M} \left(V_{\mathbf{w}}^{\pi}(\mathbf{x}_k^i) - \left[\sum_{\ell=k}^{T} \gamma^{\ell-k} r(\mathbf{u}_\ell^i, \mathbf{x}_\ell^i)\right]\right)^2 \tag{17}$$

In practice, policy gradient algorithms update the policy using a batch of trajectories (roll-outs) collected by interaction with the environment. Each trajectory is associated with a single episode, with a sample from a trajectory collected at step $k$ consisting of observation $\mathbf{o}_k$, action $\mathbf{u}_k$, and reward $r_k(\mathbf{o}_k, \mathbf{u}_k)$. Finally, gradient ascent is performed on $\theta$ and gradient descent on $\mathbf{w}$ and update equations are given by

$$\mathbf{w}^+ = \mathbf{w}^- - \beta_{\mathbf{w}} \nabla_{\mathbf{w}} L(\mathbf{w})|_{\mathbf{w}=\mathbf{w}^-} \tag{18}$$

$$\theta^+ = \theta^- + \beta_\theta \nabla_\theta J(\theta)|_{\theta=\theta^-} \tag{19}$$

where $\beta_{\mathbf{w}}$ and $\beta_\theta$ are the learning rates for the value function, $V_{\mathbf{w}}^\pi(\mathbf{o}_k)$, and policy, $\pi_\theta(\mathbf{u}_k|\mathbf{o}_k)$, respectively.

In our implementation of PPO, we adaptively scale the observations and servo both $\epsilon$ and the learning rate to target a KL divergence of 0.001.

### B. RL Problem Formulation

In this 3-DOF air-to-air missile environment, an episode terminates when the closing velocity $v_c = -\dfrac{\mathbf{r}_{\text{TM}} \cdot \mathbf{v}_{\text{TM}}}{\mathbf{r}_{\text{TM}}}$ turns negative. The agent observation $\mathbf{o}$ was given in Eq. (4) from Section II.C, and the agent action $\mathbf{u}$ is interpreted according to Eq. (4) from Section II.C. Each rollout consists of 60 episodes, and we optimize for 90,000 episodes.

The reward function maps observations and actions to a scalar reward signal $r$, and is shown below in Eqs. (20a) through (20d). Eq. (20a) penalizes the LOS curvature, encouraging the agent to curve the LOS only when it results in higher terminal rewards, as given by Eqs. (20b) and (20c), where "done" indicates the last step of an episode. We use a discount rate of 0.95 for all rewards except the terminal reward, which uses a discount rate of 0.995. Other hyperparameters are $\alpha = -0.01$, $\beta = 10$, $r_{\text{lim}} = 1$m, $\epsilon = 20$, and $\sigma_{\text{LOSC}} = 1$.

$$r_{\text{curve}} = \alpha \|\theta_{\text{LOSC}}\| \tag{20a}$$

$$r_{\text{term1}} = \begin{cases} \beta, & \text{if } \mathbf{r}_{\text{TM}} < r_{\text{lim}} \text{ and done} \\ 0, & \text{otherwise} \end{cases} \tag{20b}$$

$$r_{\text{term2}} = \begin{cases} \epsilon \exp\left(\dfrac{\|\mathbf{r}_{\text{TM}}\|^2}{\sigma_{\text{LOSC}}^2}\right), & \text{if done} \\ 0, & \text{otherwise} \end{cases} \tag{20c}$$

$$r = r_{\text{curve}} + r_{\text{term1}} + r_{\text{term2}} \tag{20d}$$

The policy and value functions are implemented using four layer neural networks with tanh activations on each hidden layer. Layer 2 for the policy and value function is a recurrent layer implemented using gated recurrent units [26]. The network architectures are as shown in Table 3, where $n_{\text{h}i}$ is the number of units in layer $i$, obs_dim is the observation dimension, and act_dim is the action dimension. The policy and value functions are periodically updated during optimization after accumulating trajectory rollouts of 60 simulated episodes.

Table 3    Policy and Value Function network architecture

| Layer | Policy Network # units | Policy Network activation | Value Network # units | Value Network activation |
|---|---|---|---|---|
| hidden 1 | $10 * \text{obs\_dim}$ | tanh | $10 * \text{obs\_dim}$ | tanh |
| hidden 2 | $\sqrt{n_{\text{h1}} * n_{\text{h3}}}$ | tanh | $\sqrt{n_{\text{h1}} * n_{\text{h3}}}$ | tanh |
| hidden 3 | $10 * \text{act\_dim}$ | tanh | 5 | tanh |
| output | act_dim | linear | 1 | linear |

## IV. Experiments

### A. RL Optimization of PN-LOSC

Optimization uses the initial conditions and vehicle parameters given in the problem formulation (Section II). The optimization history is plotted in Fig. 7, which shows the mean, mean less one standard deviation, and minimum rewards on the primary y-axis, computed over each rollout of 60 episodes. The mean and maximum number of steps per episode is plotted on the secondary y-axis.

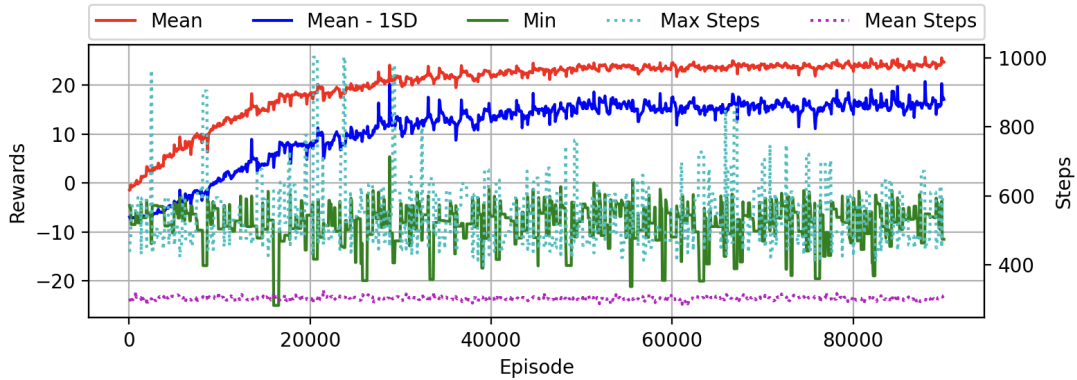

Fig. 7 Optimization Reward History

## B. Policy Testing and Comparison to Benchmarks

Here we test the optimized PN-LOSC guidance system, and compare performance to a PN and APN benchmark. Each test consists of running 5000 episodes, under different levels of maximum target acceleration capability $a_{T_{MAX}}$ and using two different target aerodynamic drag models. The randomized drag model would correspond to an unpowered maneuvering target on a ballistic trajectory, whereas the case of no target drag would correspond to a powered target such as a cruise missile holding a constant speed. Note that without target drag, the target speed is constant, and that a reduction in target speed reduces achievable acceleration (see Eq. 11a). Consequently, all three guidance systems perform better with the randomized target drag model.

The test results are given in Tables 4 and 5, and target acceleration statistics are given in Table 6. In all cases we see that the highest performance with least control effort is obtained with the PN-LOSC guidance law. Next in miss distance performance is APN, followed by PN. Importantly, APN is only optimal (in requiring the least control effort) for target step maneuvers, and we see that APN actually requires more control effort than PN, probably due to the types of target maneuvers we consider. It is worth noting that with regards to accuracy, the benefit of PN-LOSC is greatest for miss distances of less than 100cm. Finally, note that in a more realistic environment, the navigation system will provide an imperfect estimate of target acceleration. This would likely widen the performance gap between APN and PN-LOSC.

Table 4  No Target Drag, $a_{T_{MAX}} = 30g$

| | Miss (%) | | | $\|\mathbf{a}_M\|$ (m/2$^2$) | | |
|---|---|---|---|---|---|---|
| Guidance | < 100cm | < 200cm | < 300cm | $\mu$ | $\sigma$ | Max |
| PN | 52 | 77 | 85 | 31 | 33 | 336 |
| APN | 64 | 91 | 95 | 42 | 37 | 309 |
| PN-LOSC | 86 | 93 | 94 | 30 | 31 | 308 |

Sample trajectories for the PN-LOSC guidance law, with no target drag and $a_{T_{MAX}} = 30g$, are shown in Figs. 8 through 10. The left top subplot illustrates the evolution of the curvature parameter $\theta_{LOSC}$, the right top subplot plots the LOS rotation rate Eq. (6a), and the bottom sub plots show the evolution of the missile and target acceleration over the engagement. The weave maneuver plot is particularly interesting, as it appears that the LOS curvature increases as range to target decreases. By attenuating the response to the target maneuver at longer ranges, less control effort is required. Since control effort increases drag, which reduces missile speed and acceleration capability, the PN-LOSC guidance law ends the engagement (on average) with higher missile acceleration capability.

Table 5  Randomized Target Drag, $a_{T_{MAX}} = 30$g

| | Miss (%) | | | $\|\mathbf{a}_M\|$ (m/s$^2$) | | |
|---|---|---|---|---|---|---|
| Guidance | < 100cm | < 200cm | < 300cm | $\mu$ | $\sigma$ | Max |
| PN | 67 | 93 | 98 | 24 | 22 | 287 |
| APN | 71 | 97 | 100 | 34 | 30 | 334 |
| PN-LOSC | 95 | 99 | 100 | 24 | 21 | 248 |

Table 6  Target Acceleration Statistics

| $a_{T_{MAX}}$ (g) | Drag Model | $\|\mathbf{a}_T\| \mu$ (g) | $\|\mathbf{a}_T\| \sigma$ (g) | $\|\mathbf{a}_T\|$ Max (m/s$^2$) |
|---|---|---|---|---|
| 30 | None | 22 | 27 | 262 |
| 30 | Randomized | 16 | 20 | 251 |

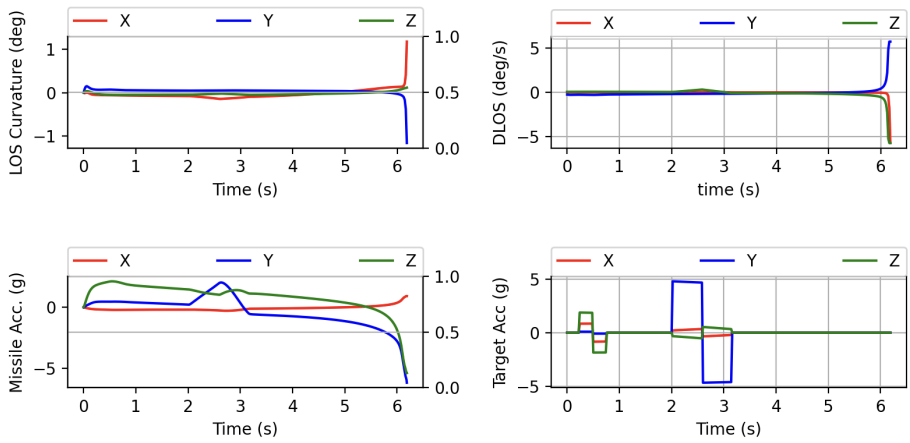

Fig. 8  Bang-Bang Maneuver

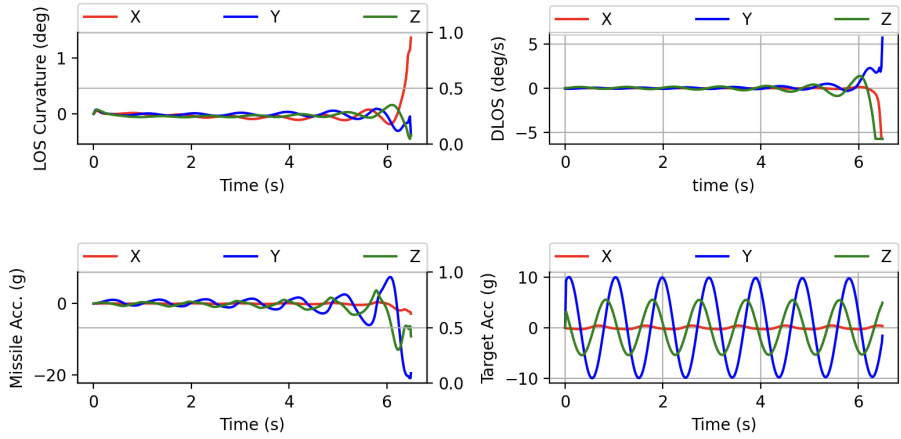

Fig. 9  Weave Maneuver

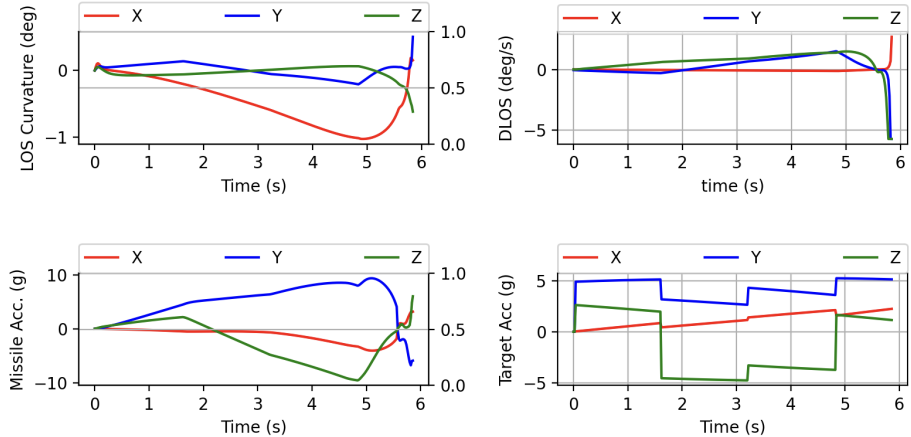

Fig. 10    Jink Maneuver

### C. Results with No Radome Refraction

As look angle dependent radome refraction results in a false indication of target motion, it is possible that at least some of the performance advantage of the PN-LOSC guidance system is attributable to increased robustness to radome refraction. The lower control effort as compared to APN supports this hypothesis, which we investigate by testing all three guidance systems with no target drag and 30g maximum target acceleration capability, but with $A_u$ and $A_v$ set to zero in the radome model (Section II.B). The results are shown in Table 7, where as expected we find that the performance of all three guidance systems improves. Importantly, we see that the improvement is greatest for the PN and APN systems, and that the accuracy of PN-LOSC is similar to that of APN, although PN-LOSC requires considerably less control effort. This suggests that the performance of the PN-LOSC guidance system is due to two factors. The first factor is an increased robustness to radome refraction, whereas the second factor is an improved ability to react to target maneuvers as compared to PN guidance. Moreover, the robustness to radome refraction is likely in at least part due to the PN-LOSC guidance system responding more intelligently to perceived target acceleration. This is supported by the lower control effort as compared to APN, and the LOS shaping and missile acceleration profile for the weave maneuver in Fig. 9. Since the parasitic attitude loop can be attenuated by increasing guidance system lag, we experimented with increasing the filter time constant to 0.05s. However, the increase in flight control response time had a negative overall impact on performance.

Table 7    No Target Drag, No Radome Refraction, $a_{T_{\text{MAX}}} = 30$g

| Guidance | Miss (%) | | | $\|\mathbf{a}_M\|$ (m/2$^2$) | | |
| --- | --- | --- | --- | --- | --- | --- |
| | < 100cm | < 200cm | < 300cm | $\mu$ | $\sigma$ | Max |
| PN | 81 | 86 | 89 | 29 | 32 | 329 |
| APN | 95 | 96 | 97 | 40 | 38 | 351 |
| PN-LOSC | 93 | 94 | 95 | 29 | 31 | 319 |

## V. Conclusion

We presented a novel method for dynamically curving the line of sight input to a guidance law using a policy optimized with reinforcement meta-learning. The PN-LOSC guidance law was then formulated by combining proportional navigation with the line of sight curvature policy. In the experiments we compared the performance of PN-LOSC to both proportional navigation and augmented proportional navigation against highly maneuvering targets, and demonstrated that PN-LOSC has improved accuracy and requires less control effort, without requiring an estimate of target acceleration. Additional experiments reveal that the performance of PN-LOSC can be attributed to two factors, the first being increased robustness to radome refraction, and the second a more intelligent response to target maneuvers. Although we demonstrated the combination of line of sight curvature with proportional navigation, the approach could also be coupled with other guidance laws. Importantly, LOS curvature could also be used to satisfy look angle and impact angle constraints, as well as constraints on load and heating rate. Future work will investigate the use of line of sight curvature in a higher fidelity six degrees-of-freedom model.